\documentclass[10pt,letterpaper]{article}

\usepackage{cogsci}

\usepackage{graphicx}

\cogscifinalcopy 

\usepackage[natbibapa]{apacite}

\DeclareUnicodeCharacter{1EF3}{y}
\usepackage{pslatex}
\usepackage{float} 

\title{Visual moral inference and communication}
 
\author{{\large \bf Warren Zhu (warrenz@cs.toronto.edu), Aida Ramezani (armzn@cs.toronto.edu)} \\
  Department of Computer Science\\ University of Toronto \\
  \AND {\large \bf Yang Xu (yangxu@cs.toronto.edu)} \\
  Department of Computer Science, Cognitive Science Program\\
  University of Toronto}

\begin{document}

\maketitle

\begin{abstract}
Humans can make moral inferences from multiple sources of input. In contrast, automated moral inference in artificial intelligence typically relies on language models with textual input. However, morality is conveyed through modalities beyond language. We present a computational framework that supports moral inference from natural images, demonstrated in two related tasks: 1) inferring human moral judgment toward visual images and 2) analyzing patterns in moral content communicated via images from public news. We find that models based on text alone cannot capture the fine-grained human moral judgment toward visual stimuli, but language-vision fusion models offer better precision in visual moral inference. Furthermore, applications of our framework to news data reveal implicit biases in news categories and geopolitical discussions. Our work creates avenues for automating visual moral inference and discovering patterns of visual moral communication in public media.


\end{abstract}

\section{Introduction}

\begin{figure}[!t]
  \includegraphics[width=\columnwidth]{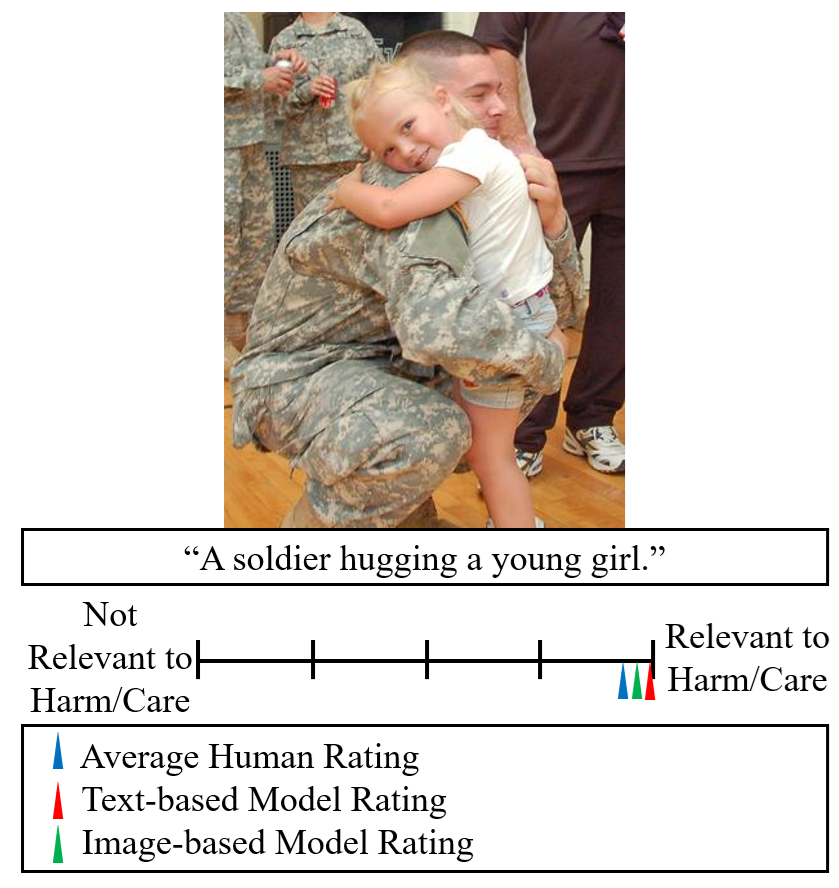}
  \caption{The photograph of a soldier hugging a child from Socio-Moral Image Database database used for analysis, accompanied by a caption generated by \textsc{Azure AI}. The text-based and image-based representations of the image and its caption are both highly related to the  moral foundation concerning Care, with the image-based representation offering a more accurate estimation of the moral content of the image based on  the ground-truth human ratings.}
  \label{fig:motivation}
\end{figure}

Morality plays a fundamental role in human cognition, but can machines tell right from wrong? This problem of computational moral
inference, or automated inference of moral values, has been a topic of
increasing relevance to artificial intelligence (AI) over the past
decade~\citep{hoover2018moral, hendrycks2021aligning,emelin-etal-2021-moral,forbes2020social,hammerl2022multilingual,Jiang2021DelphiTM,xie2020contextualized,abdulhai2023moral} pertaining to critical issues ranging from moral decision-making in autonomous vehicles~\citep{awad2018moral} to alignment of AI and humans in cultural moral norms and values~\citep{tao2024cultural,ramezani2023knowledge}. 

The dominant
approach to moral inference relies on text and considers language as the
sole medium for moral communication. For example, work from natural
language processing and computational social science has developed text-based methods for automatic
moral inference such as classifying moral sentiment~\citep{garten2016morality,johnson-goldwasser-2018-classification,johnson-goldwasser-2019-modeling,lin2018acquiring,rezapour-etal-2019-enhancing,pavan2020twitter,trager2022moral}, detecting moral sentiment change~\citep{garten2016morality,mooijman2018moralization,xie-etal-2019-text,ramezani2021unsupervised}, generating moral text generation and aligning language models with human values~\citep{hendrycks2020aligning,forbes2020social,emelin-etal-2021-moral,Jiang2021DelphiTM,ammanabrolu-etal-2022-aligning,shen-etal-2022-social}. 

This line of work has shown promise for textual moral inference. However, morality is often
conveyed through mediums beyond language. In particular, prior research shows that images (e.g., photographs) can convey moral information beyond words and facilitate human ethical reasoning~\citep{doi:10.1177/107769900608300407}. Figure~\ref{fig:motivation}
illustrates how a moral sense of care can be conveyed through our visual perception of a photograph. In this case, text-based moral inference methods may not be sufficient to capture the moral sentiment elicited by visually processing the image. As such, a quest for visual moral
inference is warranted. While existing work has proposed methods for filtering immoral or sensitive parts of an image~\citep{harminference,park2023ensuring,Jeong_2022_BMVC}, our study is distinct in that we focus on understanding how morality is communicated through the visual modality and how computational methods can automatically infer fine-grained human moral judgment in natural images at scale.

By leveraging moral information sourced from both language and vision, we contribute a fusion-based framework that extends text-based approaches to moral inference. We apply our framework to images from news media to understand how news images might communicate moral information to the public. We develop our framework to address the following two related problems:

\begin{itemize}
    \item {\bf Visual moral inference}: Can computational models drawn from AI make reliable prediction about fine-grained human moral judgment toward natural images such as photos? 
   
    \item {\bf Visual moral communication}: Can these models of visual moral inference be applied to analyzing how morals are embedded and communicated to the public through images, such as those appearing in  news articles?
   
\end{itemize}

To address the first problem, we develop supervised models for visual moral inference  using a large database of photographic images where human
moral ratings are available.\footnote{The code to replicate our framework and analysis can be found in the following repository:\\ \url{}{https://github.com/CoderWarren/Visual-Moral-Inference-and-Communication/tree/main}.} This development allows us to tackle the second
problem in an unsupervised way with minimal human intervention, which is useful because human moral ratings are
typically scarce or unavailable for public images.


In developing our  framework, we draw on the Socio-Moral Image Database (SMID)~\citep{10.1371/journal.pone.0190954}, the largest  standardized visual moral stimulus set to our knowledge. This database provides a wide range of publicly-available photographic images, rated by humans for their moral content. The database includes ratings that capture the extent to which an image is related to the moral foundations based on the Moral Foundations Theory~\citep{graham2011mapping}. 
Moral Foundations Theory  is a modern theory of morality proposing that morality is modular and depends on five 
core foundations: \emph{Care} (concerning the prevention/alleviation of suffering), \emph{Fairness} (concerning the identification of exploitation/cheating), \emph{Ingroup} (concerning self-sacrifice for the benefit of a collective), \emph{Authority} (concerning respecting/obeying superiors), and \emph{Purity} (concerning the avoidance of pathogens through the regulation of one's sexual and eating behaviors). For instance, as shown in Figure~\ref{fig:motivation}, the image of a soldier hugging a child is rated highly by human participants with respect to the Care moral foundation, while an image of a child respecting their parents would be rated highly for the Authority foundation. We demonstrate that combining visual image representations with textual representations (derived from image captions) tend to offer the most accurate prediction of fine-grained human moral ratings in this database.

Building on the visual moral inference framework, we show that it can be applied to exploring how moral information is  communicated through  images in a public news media. While prior work has  investigated how morality and moral foundations, in particular, are communicated in various forms of text-based media, including news, social media, and  child speech~\citep{fulgoni2016empirical,
ramezani2021unsupervised, shahid-etal-2020-detecting,roy-goldwasser-2020-weakly,ramezani2022emergence,hoover2020moral, trager2022moral,hofmann-etal-2022-modeling,
roy-etal-2022-towards}, how morals are communicated visually has not been comprehensively explored. Here we investigate visual moral communication   using the New York Times images during the period 2010-2018~\citep{8953244}. We find  evidence for implicit biases based on the moral content in the images displayed across different news categories and geographical regions. Here we summarize the landscape of the literature relevant to the development of our framework.




{\bf Computational approaches to moral inference.}
There has been a growing interest in using AI and computational methodologies for large-scale inference and discovery of moral values. The computational models developed under this paradigm of textual moral inference are typically grounded in established theories of human morality, and they have been applied to classify text based on different moral categories, particularly moral foundations~\citep{trager2022moral,lin2018acquiring,hoover2020moral,mooijman2018moralization,garten2016morality,johnson-goldwasser-2018-classification,johnson-goldwasser-2019-modeling,liscio-etal-2022-cross,rezapour2019moral,rezapour-etal-2019-enhancing,asprino-etal-2022-uncovering,kobbe-etal-2020-exploring,qian2021morality,pavan2020twitter, roy-goldwasser-2021-analysis,santos2019moral}. One important takeaway from this line of research suggests that people often communicate their moral views through language, and language use or language modelling can be an effective approach for making inferences about human morals at scale. 

With recent advance in generative language models, a new research area has emerged that evaluates and aligns the outputs of these models to human (moral) values. Efforts in this area vary from assessing and fine-tuning language-model generated text according to different moral theories (e.g., Virtue ethics, Utilitarianism)~\citep{hendrycks2020aligning, jin2022make,abdulhai2023moral,simmons2022moral} to grounding model outputs in crowd-sourced generated datasets of moral norms~\citep{Jiang2021DelphiTM, shen-etal-2022-social, forbes2020social,emelin-etal-2021-moral, ammanabrolu-etal-2022-aligning, liu2022aligning, bai2022constitutional}.

Our study extends the existing research on moral inference from text to the visual domain and takes an initial step toward developing fusion models for moral inference.

{\bf Psychological and cognitive studies on visual moral communication.}
Images can contain information that can go beyond and even be different from their textual description~\citep{MaC248,doi:10.1177/107769900608300407}. Research on visual moral communication suggests certain symbolic depictions of  icons and events, such as cigarettes~\citep{10.1093/hcr/hqy004}, can significantly influence people's perception of morality~\citep{DEFREITAS2018133}. For example, colors of black and white symbolize purity and pollution for a lot of people~\citep{b20df8c7-dbce-3353-b5f9-5978b1777339, BlackUniforms, doi:10.1111/j.0963-7214.2004.01502002.x,BWchild}, and certain choices of these colors can even affect people's moral judgments in other modalities such as text~\citep{ZARKADI2013355}.  

Our work builds on these existing empirical studies toward automated analysis of visual moral communication, and we demonstrate how our framework can be used for detecting implicit patterns in public communication of moral information from news images.



 {\bf Language-vision fusion models.} Besides a multitude of language models such as BERT~\citep{devlin-etal-2019-bert} and ERNIE~\citep{zhang-etal-2019-ernie}) and vision models (such as YOLO~\citep{7780460} and SAM~\citep{Kirillov2023SegmentA}, there are also various vision-language models (such as FLAVA~\citep{9880206} and BLIP~\citep{Li2022BLIPBL}), some of which have been used to automate image captioning~\citep{10.5555/3618408.3619222, 10386812}. CLIP~\citep{Radford2021LearningTV}, in particular, is a powerful vision-language model trained on 400 million image-text pairs that is capable of generating robust image and textual embeddings for downstream tasks. Although CLIP text encoders under-perform SBERT encoders~\citep{reimers-gurevych-2019-sentence} on general natural language understanding tasks, CLIP textual encoders possess a unique ability that SBERT encoders do not---they can associate a text and its visual appearance, which is more similar to human perception~\citep{chen-etal-2023-difference}. Additionally, CLIP has been used in many in state-of-the-art Vision-Language models such as DALL-E~\citep{ramesh2021zeroshot} and SAM~\citep{Kirillov2023SegmentA}. 

 For our work, we consider CLIP for visual moral inference and systematically compare it to different approaches for building a reliable computational framework that integrates images and text for reliable and automated moral inference.

\section{Data}
\subsection{Socio-Moral Image Database (SMID)}
We use the Socio-Moral Image Database~\citep{10.1371/journal.pone.0190954}. This dataset consists of 2,941 photographic images with their normative ratings (each on a 1-5 scale) for morality (``blameworthy'' to ``praiseworthy''), and relevance to the five moral foundations (``unrelated'' to ``related'') from around 2,000 human participants. By itself, SMID does not feature captions for the images, we therefore generated captions for the aforementioned images through the use of 
Microsoft’s \textsc{Azure AI} (Version 4.0, accessed through their 2024-02-01 REST API).
\footnote{The model can be found at \url{https://azure.microsoft.com/en-us/solutions/ai}}
To ensure robustness of our results, we also used \textsc{Google Vertex AI} as an alternative to \textsc{Azure AI}. Given the guardrails of \textsc{Vertex AI}, we were unable to generate captions for some sensitive images, so we therefore used \textsc{AZURE AI} for all the analyses reported here. 

\subsection{GoodNews New York Times images}
GoodNews~\citep{8953244} is a dataset of more than 466,000 New York Times images from January 2010 to June 2018, collected for image captioning. Along with images, GoodNews also provides image captions and links to their respective articles. Unlike the SMID data, images of the NYT news data had no human moral ratings. The framework we built allows one to estimate moral ratings for this news source but also flexibly for other sources of data where human ratings are not available or accessible.

\section{Computational methodology}
In this section, we provide details for the construction of our framework that comprises of visual moral inference and communication, summarized and illustrated in Figure~\ref{fig:schematic}.

\subsection{Models of visual moral inference}
We compare different representations of images in SMID to evaluate their ability to capture the moral content in images. Our analysis incorporates both textual (from image captions) and visual information, and we consider the text-based models as baselines. We summarize the different model classes as follows:

\begin{figure}[t]
  \includegraphics[width=\columnwidth]{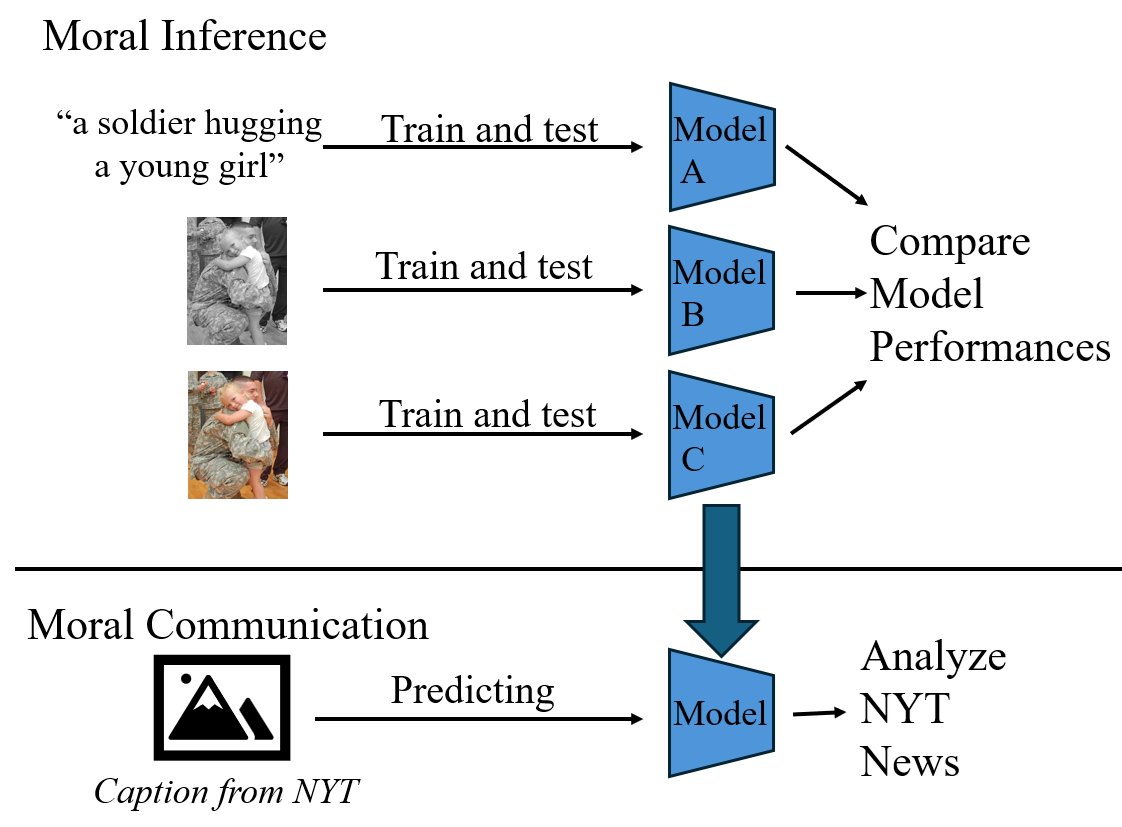}
  \caption{An illustration of our image-text fusion framework for visual moral inference and communication. Top plot: Evaluation of different text and image representations of the input figures used to train computational models for moral inference.
  Bottom plot: Applying the text-image fusion model to uncover implicit patterns of visual moral communication in news media.}
  \label{fig:schematic}
\end{figure}

\begin{itemize}
  \item {\bf Bag-of-Words (BoW):} 
We use the Bag-of-Words representation of image captions as our baseline for moral inference. Stopwords identified by \verb|NLTK| were removed. This approach captures how individual words, irrespective of their position and context, reflect human moral concerns. Although simple and less accurate than the other approaches, this baseline offers interpretability at the token level and insights into the topics that trigger the most morally relevant responses in human participants.

  \item {\bf Contextual Embedding:} We use the \verb|ViT-B/32| CLIP text-encoder~\citep{Radford2021LearningTV} and a variety of SBERT encoders~\citep{reimers-2019-sentence-bert} to generate contextual representation of image captions. The full list of SBERT encoders can be found in Table~\ref{table:final}. The CLIP text-encoder model allows us to capture how image captions are associated with visual appearances, while SBERT encoders provide better quality language representation~\citep{chen-etal-2023-difference}.
    
  \item {\bf Grayscaled Image Embedding:} 
  We produce the corresponding grayscaled representation of the images with the purpose of studying the role of color in visual moral inference. The images are then transformed into high-dimensional representations using the \verb|ViT-B/32|  CLIP image-encoder.
  
  \item {\bf Original Image Embedding:} 
    These embeddings were produced by running the original image through  \verb|ViT-B/32| CLIP image-encoder.

  \item {\bf Joint Image-Text Embedding:} 
  These embeddings were produced by adding the Original Image Embeddings to the CLIP Textual Embedding.

\end{itemize}
Our BoW vectors consisted of 1,579 dimensions, and CLIP embeddings were 512-dimensional. The size of our SBERT embeddings varied, with \verb|all-MiniLM-L6-v| and \verb|all-MiniLM-L12-v2| being 384-dimensional and the rest being 768-dimensional. Before further analysis, BoW vectors were normalized by dividing by the sum of its components, while embeddings were normalized by dividing by their Euclidean Norm.
We divided the images into an 80(training)-20(prediction) split, and used their representations to train ridge regression models for each of the dimensions in SMID related to morality (i.e., Morality, and relevance to the five moral foundations). Ridge regression involves minimizing the objective loss function: $
   ||y - XW||^2_2 + \alpha * ||W||^2_2$. Here, $y$ represents image scores we wish to predict (e.g., Morality), $X$ is the image representations (e.g., CLIP representations of the original images), and $W$ is the regression parameter. We find the best $\alpha$ (penalty parameter) using grid search on the validation set. Specifically, we run 3 rounds of 10-fold cross-validation on the training set and choose the value that yields the best $R^2$ score on average. We chose a linear regression model over more complex approaches to  compare how well the existing features (captured by the different embeddings) predict the target moral variables without further manipulation. After identifying the hyper-parameters and the most reliable representation, we trained our final model on full set of the SMID images.

\subsection{Analysis of visual moral communication}
For analyzing visual moral communication in public news images, we apply the best performing model drawn from the set of models described in the previous section on the New York Times article images as provided in the GoodNews database to obtain their estimated moral ratings. Unlike SMID images where we constructed the captions using automatic AI tools, here we use captions from the actual NYT articles. 

GoodNews further provides the publication date and the news category the articles fall into. From these categories we extracted images of different regions ({\it nyregion}, {\it us}, {\it world/europe}, {\it world/asia}, {\it world/africa}, and {\it world/middleeast}) and article topics ({\it business}, {\it health}, {\it sports}, {\it technology}, and {\it science}) for exploring differences and potential biases in moral communication under these  categories.

\begin{table*}
\centering
\begin{tabular}{||l | c c c c c c | c||}
 \hline
 Model & M & A(R) & F(R) & C(R) & I(R) & P(R) & Average\\
 \hline
  \hline
BoW  & 0.3747 & 0.4441 & 0.3049 & 0.3871 & 0.4146 & 0.3368 & 0.3770\\
\hline
SBERT {\tiny(\verb|all-mpnet-base-v2|)} & \underline{0.4520} & 0.5011 & 0.3475 & 0.4550 & 0.4580 & 0.3628 & 0.4294\\
\hline
SBERT {\tiny(\verb|multi-qa-mpnet-base-dot-v1|)} & 0.4244 & 0.4932 & \underline{0.3695} & 0.4483 & 0.4496 & 0.3417 & 0.4211\\
\hline
SBERT {\tiny(\verb|all-distilroberta-v1|)} & 0.4374 & 0.4973 & 0.3469 & 0.4588 & 0.4552 & 0.3371 & 0.4221\\
\hline
SBERT {\tiny(\verb|all-MiniLM-L12-v2|)} & 0.4276 & 0.4862 & 0.3435 & 0.4489 & \underline{0.4700} & 0.3397 & 0.4193\\
\hline
SBERT {\tiny(\verb|multi-qa-distilbert-cos-v1|)} & 0.4281 & 0.5083 & 0.3529 & 0.4351 & 0.4544 & \underline{0.3661} & 0.4242\\
\hline
SBERT {\tiny(\verb|all-MiniLM-L6-v2|)} & 0.4113 & 0.4806 & 0.3385 & 0.4253 & 0.4481 & 0.3405 & 0.4074\\
\hline
CLIP {\tiny(Caption)} & 0.4306 & \underline{0.5172} & 0.3519 & \underline{0.4677} & 0.4696 & 0.3627 & \underline{0.4333}\\
\hline
\hline
CLIP {\tiny(Grayscaled Image)}& 0.6245 & 0.6214 & 0.5416 & 0.5875 & 0.6095 & 0.5338 & 0.5864\\
\hline
CLIP {\tiny(Colour Image)} & 0.6742 & 0.6560 & {\bf 0.5796} & 0.6360 & 0.6408 & 0.5757 & 0.6270\\
\hline
CLIP {\tiny(Joint)} & {\bf 0.6755} & {\bf 0.6701} & 0.5704 & {\bf 0.6513} & {\bf 0.6427} & {\bf 0.5822} & {\bf 0.6320}\\
\hline
\end{tabular}
\caption{\label{table:final} The $R^2$ scores between predicted and actual ratings of the SMID test set for the best performing models. Purely text-based models are found on the top section of the table--models involving images in some way are found on the bottom section. Here, M, A(R), F(R), C(R), I(R), and P(R) respectively stand for Morality, and relevance to Authority, Fairness, Care, Ingroup and Purity. The best $R^2$ scores for each predicted variable are shown in \textbf{bold}. The best scores using the text-based models are \underline{underlined}.} 
\end{table*}

\section{Results}
\subsection{Evaluating models of visual moral inference}

The overall performance of the models after hyperparameter optimization are summarized in Table~\ref{table:final}. 
The joint image-text embedding produces the most reliable representation for predicting moral ratings with the average $R^2 = 0.6320$. This is closely followed by the original image representation with the average $R^2 = 0.6270$. Morality and the relevance to the Authority, Care, and Ingroup moral foundations were predicted more accurately than Fairness and Purity moral foundations by all models. Moreover, the grayscaled image representations performed slightly worse across all 6 variables of interest compared to our coloured image models, suggesting that similar to humans \citep{ZARKADI2013355}, colour also acts as a visual moral stimuli in CLIP representations. In all cases, the image-based representation produced by CLIP outperformed the text-based models, with the worst-performing visual model (grayscaled image representations) having a correlation of at least $0.10$ higher for each of the 6 variables of interest compared to the best performing text-based model (CLIP-embedding textual model). 

Among the text-based models, CLIP text representation slightly outperforms each of the SBERT model. While SBERT is shown to be better at natural language understanding tasks, our goal primarily revolves around replicating human ratings for images, which leverages CLIP's effectiveness in associating a text with its visual appearance \citep{chen-etal-2023-difference}.
Despite the comparatively poor performance of the BoW model, we found that it was capable of detecting concepts and actions related to moral foundations, such as {\it soldier}, {\it police} and {\it saluting} for Authority and {\it crashed}, {\it destroyed}, and {\it explosion} for Care. 

Our results show that the visual stimuli in these photographic images elicit moral responses from human participants that cannot be accurately predicted by relying solely on textual representations of images. Consequently, models that rely only on captions cannot reliably infer the actual moral ratings elicited by such images in participants. In summary, given that the CLIP model itself is developed with joint image-text training, our results demonstrate the effectiveness of incorporating  visual stimuli with  textual information for both moral and non-moral (i.e., valence and arousal)  prediction based on fine-grained human judgment.

\subsection{Visual moral communication in public news}
To investigate how moral values may be embedded and communicated through visual stimuli in public media, we use our best performing model (i.e., joint image-text model) on the article images in the New York Times articles published in 2010-2018, using the GoodNews dataset~\citep{8953244}. 

\begin{figure*}[!h]
  \centering
  \includegraphics[width=.8\textwidth]{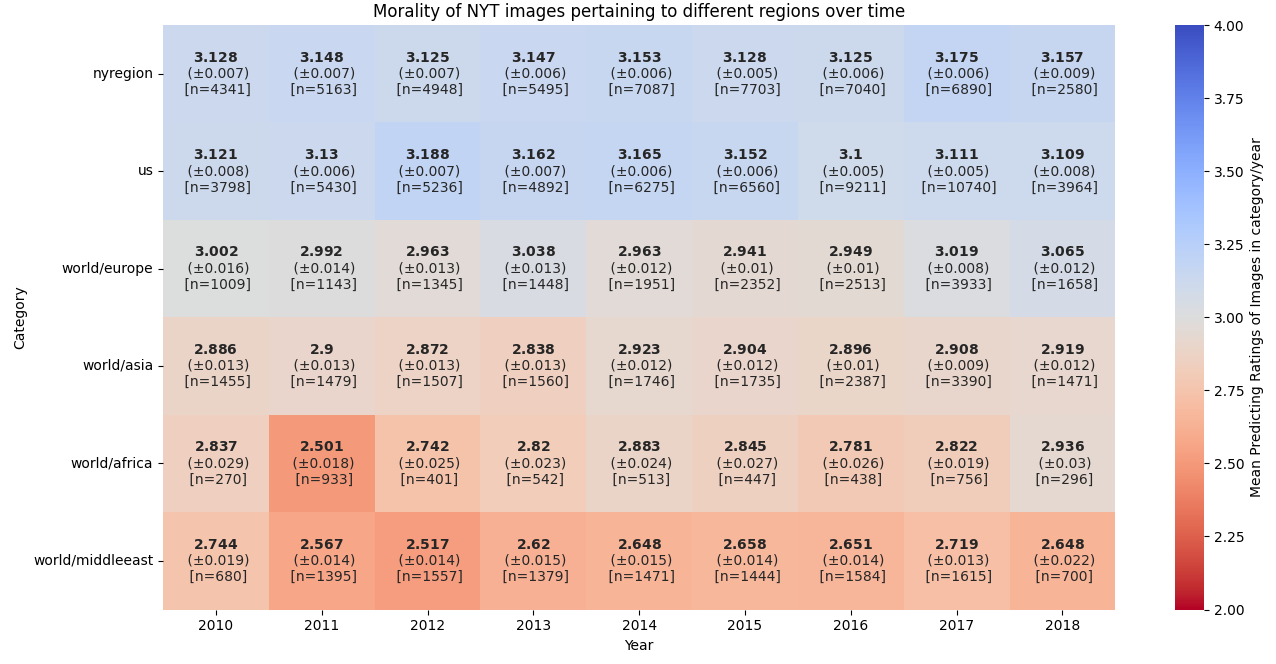}
  \caption{The predicted Morality scores of images corresponding to each regional category. Within each cell, the mean Morality score has been \textbf{bolded} on top, the standard errors of each mean are within the parentheses in the middle, and the number of images matching each year/category are in the square brackets on the bottom.}
  \label{figure:regional bias}
\end{figure*}

{\textbf{Public news images contain implicit moral biases across regions.}} We find evidence for moral bias from images included in NYT news categorized across different regions: {\it nyregion}, {\it us}, {\it world/europe}, {\it world/asia}, {\it world/africa} and {\it world/middleeast}. As shown in Figure~\ref{figure:regional bias}, images in {\it nyregion} (region of New York) and {\it us} (USA) exhibit more morally positive (i.e., moral) information and sentiment compared to the images in {\it world/europe}, {\it world/asia}, {\it world/africa}, and {\it world/middleeast}. Moreover, images in the {\it world/africa} and  {\it world/middleeast} regions are associated more strongly with Care and Purity moral foundations compared to the other regions, as most NYT images associated with {\it world/africa} and  {\it world/middleeast} typically depict warfare and the aftermath of combat. Examples of captions of these images include ``Children receiving treatment after the gas attack'' and ``Rebels with the body of the commander''.



\begin{figure*}[!h]
  \centering
  \includegraphics[width=.8\textwidth]{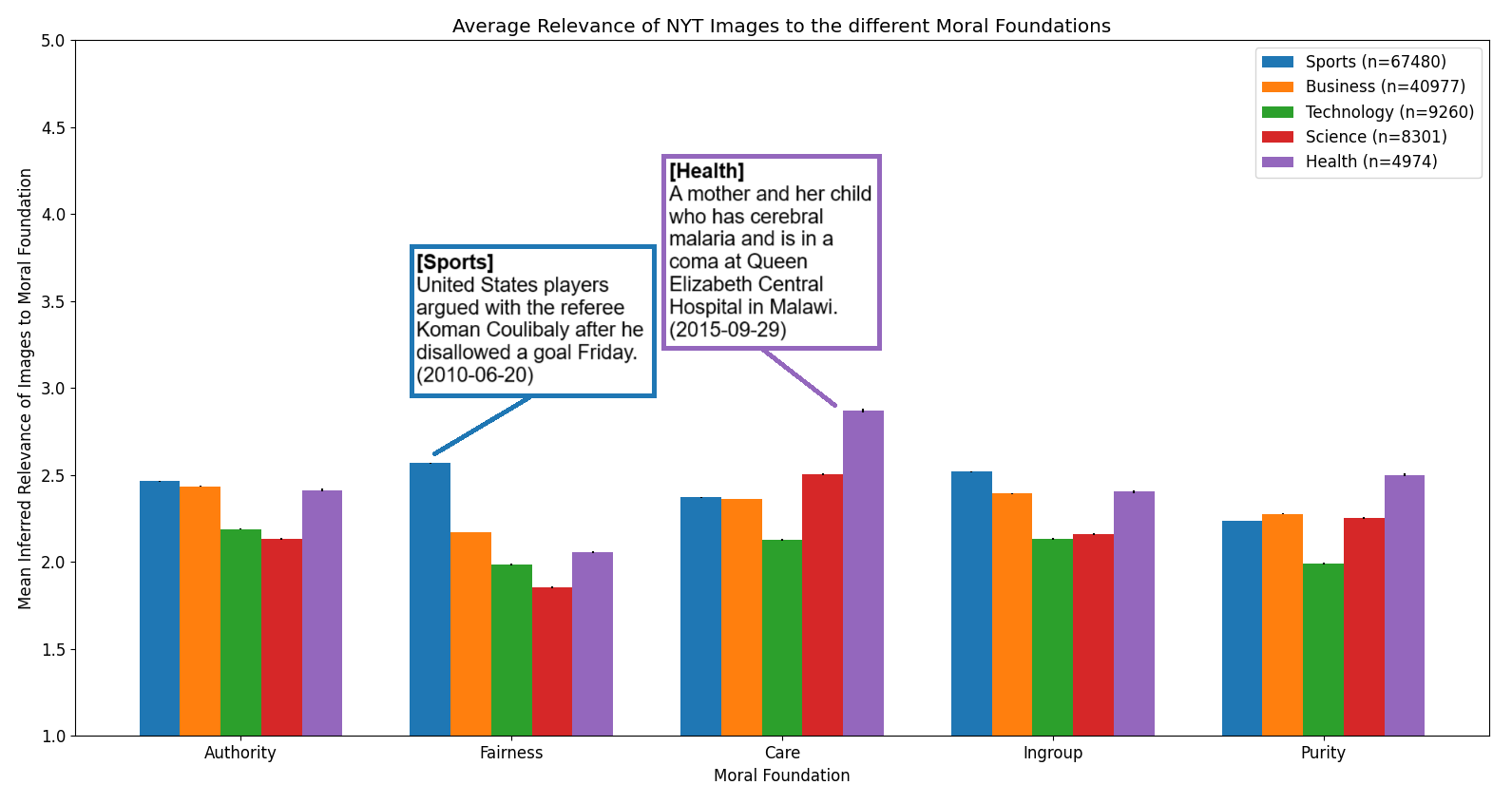}
  \caption{\label{fig: Category Averages} The mean predicted relevance to different moral foundations across all years for the news categories of interest. A rating of 1 indicates that an image is unrelated to the moral foundation, while a rating of 5 indicates that an image is highly related to the moral foundation. The number of images in each category can be found in the top right corner. Error bars indicate the standard error of the mean. Captions of sample images with high moral relevance are shown above the corresponding bars.}
\end{figure*}

{\textbf{Different news categories engage different moral foundations.}}
Table~\ref{CategoryMoralAverages} displays the average predicted morality for news categories {\it health}, {\it sports}, {\it business}, {\it science}, and {\it technology}. {\it Health} has the highest average Morality score with many images of people and animals being treated in clinical settings. 

\begin{table}[h!]
\centering
\begin{tabular}{|l|c|} 
 \hline
 Category & Morality \\
 \hline\hline
 \textbf{{\it health}} & \textbf{3.2821 ($\pm$ 0.0069)}\\
 \hline
{\it sports} & 3.1177 ($\pm$ 0.0015)\\
\hline
{\it business} & 3.0967 ($\pm$ 0.0020)\\
\hline
{\it science} & 3.0884 ($\pm$ 0.0050)\\
\hline
{\it technology} & 3.0880 ($\pm$ 0.0038)\\
\hline

\end{tabular}
\caption{\label{CategoryMoralAverages}The mean predicted Morality scores for each of our categories, along with the standard errors of the mean.}
\end{table}

Furthermore, Figure~\ref{fig: Category Averages} displays the average relevance to each of the different moral foundations moral scores.
We find that the five categories exhibit different moral foundations averaged across all years. For example, {\it sports}, with many images of referees, engages the Fairness moral foundation more so than other categories. This category is also associated with the Ingroup moral foundation, with images showing sports fans, teammates, and families banding together. 
Among these categories, images in {\it health} appeal the most to the Care moral foundation. This is followed by {\it science}, which also includes images depicting clinical treatment. {\it Health} also consists of many images of people praying, possibly making it most relevant to the Purity moral foundation across all five categories.


Bootstrap testing was conducted to verify that these results are not based on random chance. We found strong evidence against {\it health} having the highest average Morality score ($P < 0.01$) and relevance to the Care ($P<0.0001$) and Purity ($P < 0.001$) simply due to random chance. Evidence was also found against {\it sports} having the highest relevance to Fairness ($P<0.0001$) simply due to chance. 


\section{Discussion and conclusion}

Automated moral inference is an area of increasing relevance to AI and it has a foreseeable impact on empirical research in morality.  We show that moral inference solely based on text is inadequate for understanding the diverse mediums through which morals may be communicated. We  found evidence that text can strengthen moral inference from visual stimuli, but text derived from images alone is not sufficient to capture the fine-grained moral sentiment toward images. Previous research has shown that visual perception can influence one's moral interpretation of text \citep{ZARKADI2013355, doi:10.1177/107769900608300407}, and we have fulfilled a need to developing methods for reconstructing fine-grained human moral judgment of natural images. We also demonstrated the utility of our framework to identifying implicit biases in images taken from public news. Future work may extend our framework to accommodate other sources of information such as sound and audios, movies and videos, and therefore exploring how different modalities may reflect or affect moral judgment. We hope that our work will generate
further research in integrating language and vision with other modalities toward a more holistic and multimodal approach to  moral
inference.

\clearpage

\section*{Acknowledgments}

This research is supported by an Ontario Early Researcher Award \#ER19-15-050 to YX.



\bibliographystyle{apacite}
\bibliography{CogSci_Template.bib}

\end{document}